\documentclass[journal]{IEEEtran}
\usepackage{cite}
\usepackage{amsmath,amssymb,amsfonts}
\usepackage{algorithmic}
\usepackage{graphicx}
\usepackage{textcomp}
 \usepackage[colorlinks=true, citecolor=blue, urlcolor=blue, linkcolor=black]{hyperref}
\def\BibTeX{{\rm B\kern-.05em{\sc i\kern-.025em b}\kern-.08em
    T\kern-.1667em\lower.7ex\hbox{E}\kern-.125emX}}
    
\usepackage{color}    
\usepackage[normalem]{ulem}
\usepackage{soul}

\begin{document}


\title{LOTR: Face Landmark Localization Using Localization Transformer}




\markboth
{Accepted for publication in IEEE Access, vol.10, 2022 but has not been fully edited. DOI: 10.1109/ACCESS.2022.3149380}
{Accepted for publication in IEEE Access, vol.10, 2022 but has not been fully edited. DOI: 10.1109/ACCESS.2022.3149380}

\author{\IEEEauthorblockN{
        Ukrit Watchareeruetai\IEEEauthorrefmark{1},
        Benjaphan Sommana\IEEEauthorrefmark{1},
        Sanjana Jain\IEEEauthorrefmark{1},
        Pavit Noinongyao\IEEEauthorrefmark{1}\IEEEauthorrefmark{2},
        Ankush Ganguly\IEEEauthorrefmark{1},
        Aubin Samacoits\IEEEauthorrefmark{1},
        Samuel W.F. Earp\IEEEauthorrefmark{1}\IEEEauthorrefmark{2},
        and Nakarin Sritrakool\IEEEauthorrefmark{3}\\
        }
\IEEEauthorblockA{\IEEEauthorrefmark{1}Sertis Vision Lab, Sukhumvit Road, Watthana, Bangkok, 10110, Thailand\\}
\IEEEauthorblockA{\IEEEauthorrefmark{2}QIS Capital, Sukhumvit Road, Watthana, Bangkok, 10110, Thailand\\}
\IEEEauthorblockA{\IEEEauthorrefmark{3}Department of Mathematics and Computer Science, Faculty of Science, Chulalongkorn University\\Phayathai Road, Pathum Wan, Bangkok 10330, Thailand}
\thanks{E-mail: uwatc@sertiscorp.com}
\thanks{Nakarin Sritrakool has contributed to this work during his internship at Sertis Vision Lab.m}
}

\maketitle

\begin{abstract}
This paper presents a novel Transformer-based facial landmark localization network named Localization Transformer (LOTR). The proposed framework is a direct coordinate regression approach leveraging a Transformer network to better utilize the spatial information in a feature map. An LOTR model consists of three main modules: 1) a visual backbone that converts an input image into a feature map, 2) a Transformer module that improves the feature representation from the visual backbone, and 3) a landmark prediction head that directly predicts landmark coordinates from the Transformer's representation. Given cropped-and-aligned face images, the proposed LOTR can be trained end-to-end without requiring any post-processing steps. This paper also introduces a loss function named smooth-Wing loss, which addresses the gradient discontinuity of the Wing loss, leading to better convergence than standard loss functions such as L1, L2, and Wing loss. Experimental results on the JD landmark dataset provided by the First Grand Challenge of 106-Point Facial Landmark Localization indicate the superiority of LOTR over the existing methods on the leaderboard and two recent heatmap-based approaches. On the WFLW dataset, the proposed LOTR framework demonstrates promising results compared with several state-of-the-art methods. Additionally, we report an improvement in the performance of state-of-the-art face recognition systems when using our proposed LOTRs for face alignment.
\end{abstract}




\section{Introduction}\label{sec:intro}

\begin{figure*}
    \centering
    \includegraphics[width=0.9\textwidth]{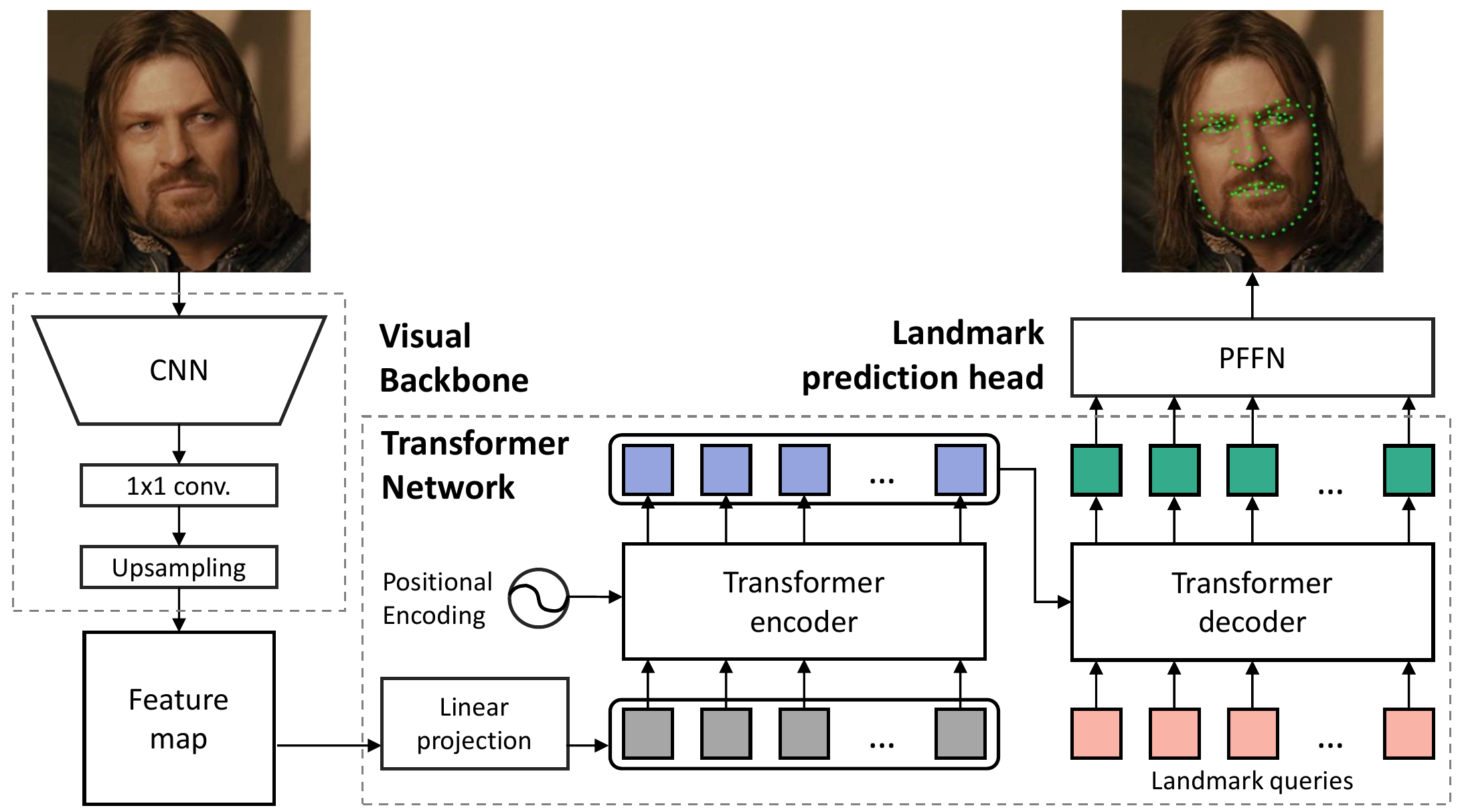}
    \caption{The overview of Localization Transformer (LOTR). It consists of three main modules: 1) a visual backbone, 2) a Transformer network, and 3) a landmark prediction head.}
    \label{fig:lotr}
\end{figure*}

Landmark localization focuses on estimating the position of each predefined key point in an image.
For face landmark localization, these key points represent different attributes of a human face, e.g., the contours of the face, eyes, nose, mouth, and eyebrows.
Over the past decade, face recognition systems have leveraged these landmarks for alignment, making face landmark localization an intrinsic part of these systems
\cite{earp2021sub, kumar2009,wolf2009,parkhi2015,schroff2015,LiuWeiyang2017,wang2018,deng2018,an2020}.
Apart from face alignment (e.g., \cite{Barra2018, Kazemi2014}), face landmark localization also aids in solving problems like face animation \cite{cao2013}, 3D face reconstruction \cite{Roth2015,Dou2017,Feng2018}, synthesized face detection \cite{yang2019}, emotion classification \cite{day2016,munasinghe22018}, and facial action unit detection \cite{hinduja2020}.
Although facial landmark localization is a substantial area of research in computer vision, given its wide range of applications, it is a challenging task owing to its dependency on variations in face pose, illumination, and occlusion \cite{Dibeklioglu2008, earp2021sub}.

Since its formulation, various statistical approaches have been proposed, such as Active Shape Model (ASM) \cite{Cootes1995} and Active Appearance Model (AAM) \cite{Cootes1998}, to solve the face landmark localization problem.
These models take in prior information about a face (e.g., the face shape or texture) and subsequently fine-tune the model parameters from the provided face image.
In addition, research has also been done in training patch-based detectors and component detectors to predict each landmark on local patches and anatomical components on a face image, respectively \cite{Liang2008, Zhu2012, Amberg2011, Belhumeur2013, Efraty2011}. 
However, due to the lack of global contextual information, the landmark configurations from these approaches are constrained on the face shape.

Since the early 2010s, different variants of Convolutional Neural Networks (CNNs) have been developed, as an alternative to the aforementioned approaches, due to their ability to extract contextual information from an image.
Two approaches, namely coordinate regression and heatmap regression, have been widely adopted with a CNN variant as a backbone.

In coordinate regression, fully-connected layers are added at the end of a CNN to predict each landmark's coordinates.  
Notable works include \cite{Sun2013, Zhou2013}, and \cite{Zhang2014} which proposed multi-level cascaded CNNs to localize facial landmarks.
Combining coordinate regression with multi-level cascaded CNNs has been instrumental in predicting the facial landmarks.
In these frameworks, the early levels aim to learn about a rough estimate of the landmarks, while the deeper levels seek to fine-tune the prediction.
The main caveat of these approaches is the high architectural complexity, thereby increasing inference time.
To address this complexity issue, \cite{ZhangLuo2016} and \cite{Ranjan2016} utilized multi-task learning to obtain lighter yet more robust models.
However, most regression-based approaches suffer from spatial information loss due to the compression of feature maps before the fully-connected layers \cite{earp2021sub}.

Recently, heatmap-based approaches (e.g., \cite{Kowalski2017, xiong2020, Mahpod2018, earp2021sub}) have extensively been used for face landmark localization as they better utilize spatial information to boost the performance compared with coordinate regression methods.
These methods predict spatial probability maps wherein each pixel is associated with the likelihood of the presence of a landmark location.
Additionally, these approaches lead to better convergence than coordinate regression techniques and achieve state-of-the-art performance on several benchmark datasets.
However, these approaches usually rely on a computationally intense post-processing step to convert heatmaps into predicted landmarks, resulting in an increase in the inference time \cite{earp2021sub}. 
Therefore, tackling this issue is the primary motivation behind this paper.

Although the coordinate regression methods mentioned so far may suffer from the issue of spatial information loss, they offer an end-to-end solution at a lower computational complexity than heatmap-based approaches. 
In this work, we re-investigate the direct coordinate regression approach for facial landmark localization but exploit a more sophisticated neural network architecture, i.e., the Transformers \cite{Vaswani}, to address the issue of spatial information loss.
A Transformer network is a sequence transduction model comprising an encoder-decoder architecture that utilizes attention mechanisms.
With the scaling successes in natural language processing (NLP) achieved by Transformers, researchers have developed different variations of the Transformer framework for computer vision tasks.
One such variant is Detection Transformer (DETR) \cite{Carion2020} which performs objection detection.
Inspired by DETR, we propose a Transformer-based facial landmark localization network named Localization Transformer (LOTR). 
As shown in Fig. \ref{fig:lotr}, LOTR consists of three main modules: 1) a visual backbone, 2) a Transformer network, and 3) a landmark prediction head. 
Firstly, LOTR adopts a CNN with optional upsampling layers to convert an input image into a feature map and reshapes it to a sequence of tokens, each representing a pixel in the feature map. 
The Transformer module then accepts this feature sequence and a fixed-length landmark queries as input, and produces a sequence of tokens as output.
Finally, the prediction head, which is a Position-wise Feed-Forward Network (PFFN), transforms each token into its corresponding landmark's coordinates. 
Given cropped-and-aligned face images as input, the proposed LOTR can be trained end-to-end without requiring any post-processing step.

The key contributions of this research are summarized as follows:
\begin{itemize}
\item We propose a Transformer-based landmark localization network named Localization Transformer (LOTR).
Unlike the heatmap-based approaches, LOTR does not require any additional post-processing or heatmap representation, reducing computational complexity and yielding a more efficient network.
To the best of our knowledge, this is the first research investigating the use of Transformers in the direct regression of landmarks.

\item We demonstrate that the proposed LOTR framework detects facial landmarks accurately.  
Experimental results indicate the superiority of the proposed LOTR over other algorithms on the leaderboard of the First JD-landmark localization challenge and two recent heatmap-based methods \cite{earp2021sub, xiong2020}.
On another benchmark, WFLW dataset \cite{Wu2018}, the results show the proposed LOTR method is comparable with several state-of-the-art methods.

\item We further analyze the model size and inference time of the different variants of the LOTR framework. 
Compared with a heatmap-based method \cite{earp2021sub} and a CNN-based direct coordinate regression method, the proposed LOTR outperforms both  of these methods in terms of prediction accuracy, model size, and computational complexity.

\item We also investigate the effect of standard loss functions on model training and propose a modified loss function, namely smooth-Wing loss, which addresses gradient discontinuity and training stability issues in an existing loss function called the Wing loss \cite{Feng2017}. 
Experimental results show improved performance for the LOTR models trained with the proposed smooth-Wing loss.

\item We report an improvement in state-of-the-art face recognition performance on several benchmark datasets, such as CFP-FP \cite{cfp-fp}, CPLFW \cite{cplfw}, IJB-B \cite{ijbb}, and IJB-C \cite{ijbc}, using five naive landmarks extracted from the predictions of our proposed LOTR models for face alignment.

\end{itemize}

The remaining of this paper is organized as follows:
Section \ref{sec:related} provides more detail of related work.
Section \ref{sec:proposed} presents the proposed method, i.e., LOTR.
Section \ref{sec:exp} explains how experiments were setup and discusses the results.
Section \ref{sec:conclusion} concludes the paper.


\section{Related Work}\label{sec:related}

\subsection{Direct coordinate regression}\label{subsec:direct}
As discussed in Section \ref{sec:intro}, direct coordinate regression was widely adopted to solve landmark localization problems in the early research period.
Regression-based approaches \cite{Sun2013, Zhang2014, zhang2014facial, ZhangLuo2016, Ranjan2016, Feng2017, Xuanyi2018} typically use a CNN along with a dense layer at the end to predict the landmark locations. 
Usually, the choice of loss for coordinate regression is either mean-absolute error (L1 loss) or mean-squared error (L2 loss).
Recently, Feng \textit{et al.} \cite{Feng2017} introduced a new loss function called Wing loss for robust facial landmark localization using regression, discussed in Section \ref{subsec:wing}.
Furthermore, Dong \textit{et al.} \cite{Xuanyi2018} trained a facial landmark localization model with style-aggregated images from a generative adversarial module along with the original images to increase the robustness of the variance of image styles.

Despite the advancement in regression-based techniques for landmark localization, spatial information loss remains one major drawback for this approach.
This research addresses this limitation by inserting a Transformer network between a CNN backbone and the fully-connected layers to preserve spatial information by leveraging a positional encoding, discussed in Section \ref{sec:proposed}.

\subsection{Heatmap regression}\label{subsec:heat}
As an alternative to coordinate regression methods, heatmap regression approaches tackle the spatial information loss issue by generating spatial probability maps. 
A spatial probability map is a heatmap with pixel values corresponding to the probability of a landmark being in a certain location.
Recent heatmap regression approaches (e.g., \cite{Kowalski2017, Mahpod2018, Wu2018, Sun2019HighResolutionRF, Kumar2020}) have demonstrated state-of-the-art performance on facial landmark localization. 

A common practice for generating a ground-truth heatmap for each landmark is to compute a probability map (e.g., \cite{Newell2016}, \cite{Wei2016}) by fitting a bivariate Gaussian function with an offset relative to the landmark, which is defined as:
\begin{equation}
    \label{eq:generic_heatmap}
    \text G(x, y) =
        \ \mbox{exp}\Big[-{\frac{(x - x_{0})^2 + (y - y_{0})^2}{2{\sigma}^2}}\Big] ,
\end{equation}
where ($x$, $y$) is any location on the heatmap and ($x_{0}$, $y_{0}$) represents the ground-truth coordinates of a landmark. 
A simple approach to obtain a predicted landmark location from a heatmap during inference is using the argmax operation; however, the heatmap's resolution becomes a limiting factor. 
Recent approaches have proposed efficient ways to obtain sub-pixel localization \cite{Xiao2018SimpleBF, zhang2019distributionaware}.
For instance, Zhang \textit{et al.} \cite{zhang2019distributionaware} proposed to fit a heatmap with the following formulation to obtain a predicted landmark location:
\begin{equation}
    \label{eq:guassian_fitting}
    \text G(\mathbf{x}, \mathbf{\mu}, \mathbf{\Sigma}) =
        \ \frac{\text{exp}(- \frac{1}{2}(\mathbf{x} - \mathbf{\mu})^{T}\mathbf{\Sigma}^{-1} (\mathbf{x} - \mathbf{\mu}))}{2 \pi |\mathbf{\Sigma}|^{\frac{1}{2}}} ,
\end{equation}
where $\mathbf{x}=(x, y)$ is any location on the heatmap, $\mathbf{\mu}$ represents an estimated location of the landmark, and $\mathbf{\Sigma}$ is the co-variance matrix between the $x$ and $y$ coordinates.

Earp \textit{et al.} \cite{earp2021sub} exploited this sub-pixel inference with intermittent shuffling of upsampling layers and a bag of tricks to achieve the second rank on the JD-landmark-2 validation set \cite{liu2019}.
While successfully preserving spatial information and achieving state-of-the-art performance on facial landmark localization, heatmap regression methods suffer from high computational post-processing complexity in converting heatmaps into landmark coordinates.

Xiong \textit{et al.} \cite{xiong2020} proposed a vectorization approach in which each set of vector labels represents a ground-truth landmark's position using a quasi-Gaussian distribution.
This proposed distribution is a Gaussian density function enhancing the distribution peak with an additional constant $\theta$ while being bounded by a threshold $3\sigma$. 
This method also converts a predicted heatmap into a set of vectors that encode spatial information.
Although this vectorization technique leads to a reduction in post-processing complexity, like other heatmap regression methods, this approach also hinders achieving an end-to-end pipeline due to a post-processing step during inference.

\subsection{Transformer-based heatmap approach}\label{subsec:transformer_heatmap}

Lan \textit{et al.} \cite{lan2021hih} proposed an approach that combines a Transformer encoder with heatmap regression.
They mainly focused on reducing the quantization error that occurs from down-sampling operations. 
The authors named the proposed architecture Heatmap in Heatmap (HIH), which takes advantage of two heatmap categories: integer and decimal heatmaps.
An integer heatmap is a probability map extracted from a CNN backbone, e.g., Stacked Hourglass Network \cite{Newell2016}, which provides a rough estimate of a landmark's location.
On the other hand, a decimal heatmap gives a more fine-grained offset prediction. 
For each landmark location, they fitted a decimal heatmap with a bivariate Gaussian density function as in \eqref{eq:generic_heatmap}.
During inference, the final landmark locations, in both integer and decimal heatmaps, are computed from the maximum probability coordinates.
The authors conducted experiments comparing two different architectures, namely a CNN and a Transformer encoder in combination with a CNN backbone.
The proposed method outperforms the state-of-the-art algorithms (e.g., \cite{Wu2018, Wang2019c, Kumar2020, Feng2017}) on WFLW-full \cite{Wu2018} and COFW \cite{Burgos2013} datasets.
They also reported that the HIH approach with a CNN yields more precise landmarks than a Transformer encoder.

Even though Lan \textit{et al.} \cite{lan2021hih} are the first who employ a Transformer encoder with heatmap regression to tackle facial localization problems, the main caveats of the heatmap regression approach, namely the post-processing complexity and the lack of an end-to-end pipeline, remain unaddressed.

\subsection{DETR}\label{subsec:detr}
Carion \textit{et al.} \cite{Carion2020} proposed a Transformer \cite{Vaswani} framework for end-to-end object detection.
Their proposed Detection Transformer (DETR) framework achieves comparative performance to Faster R-CNN \cite{girshick2014rich} on the COCO dataset \cite{lin2014}.
The DETR framework performs the task of object detection by combining a CNN architecture with a Transformer.
It exploits a pre-trained CNN backbone to extract a low-resolution feature map from an input image.
This feature map is then converted into a sequence and fed to an encoder model, consisting of a Multi-Head Attention (MHA) module and a PFFN.
The MHA module aims to find the relationship between input sequence tokens with each head computing "attention" by linearly projecting each token into query, key, and value vectors.
Let $\mathbf{Q} \in \mathbb{R}^{N_q \times D}$ denote a query sequence consisting of $N_q$ tokens of dimension $D$, while $\mathbf{K}$ and $\mathbf{V} \in \mathbb{R}^{N_{kv} \times D}$ denote a key and value sequences of length $N_{kv}$, respectively. 
The MHA is defined as follows:
\begin{equation}
    \text{MHA}(\mathbf{Q}, \mathbf{K}, \mathbf{V}) = \text{concat}(\mathbf{H}_1, ..., \mathbf{H}_{M})\mathbf{W}_i^o,
\end{equation}
\begin{equation}
    \mathbf{H}_i =  \text{attention}(\mathbf{Q}\mathbf{W}_i^q, \mathbf{K}\mathbf{W}_i^k, \mathbf{V}\mathbf{W}_i^v),
\end{equation}
\begin{equation}
    \text{attention}(\mathbf{Q}, \mathbf{K}, \mathbf{V}) = \text{softmax}(\frac{\mathbf{Q}\mathbf{K}^\top}{\sqrt{D^{\prime}}})\mathbf{V},
\end{equation}
where $M$ is the number of heads; $\mathbf{W}_i^q$, $\mathbf{W}_i^k$, $\mathbf{W}_i^v \in \mathbb{R}^{D \times D^{\prime}}$, and $\mathbf{W}_i^o \in \mathbb{R}^{D \times D}$ are learnable projection matrices; and $D^{\prime} = D/M$.
Fixed positional encoding, which encodes the spatial positions of the learned features, is added to the input of each attention layer \cite{Carion2020}.
The embeddings generated by the encoder are then fed to the decoder network, which also uses MHA mechanisms \cite{Carion2020}.
The decoder model decodes the embeddings in parallel at each decoder layer.
The decoder output is then fed into a fixed number of PFFNs, also known as the prediction heads, to generate a set of class predictions and predicted bounding boxes.
The DETR model predicts all objects at once, computed in parallel.
The authors trained the model end-to-end using Hungarian loss to compare sets, performing bipartite matching between predicted and ground-truth objects.

Inspired by DETR, we propose the Localization Transformer (LOTR) to predict landmarks for the facial landmark localization task.
The main differences between the two are: 1) LOTR does not have a class prediction head since all predicted landmarks belong to single object class, and 2) LOTR does not require the Hungarian loss since the number of landmarks is fixed.

\subsection{Wing loss}\label{subsec:wing}

\begin{figure*}
    \centering
    \includegraphics[width=1.0\textwidth]{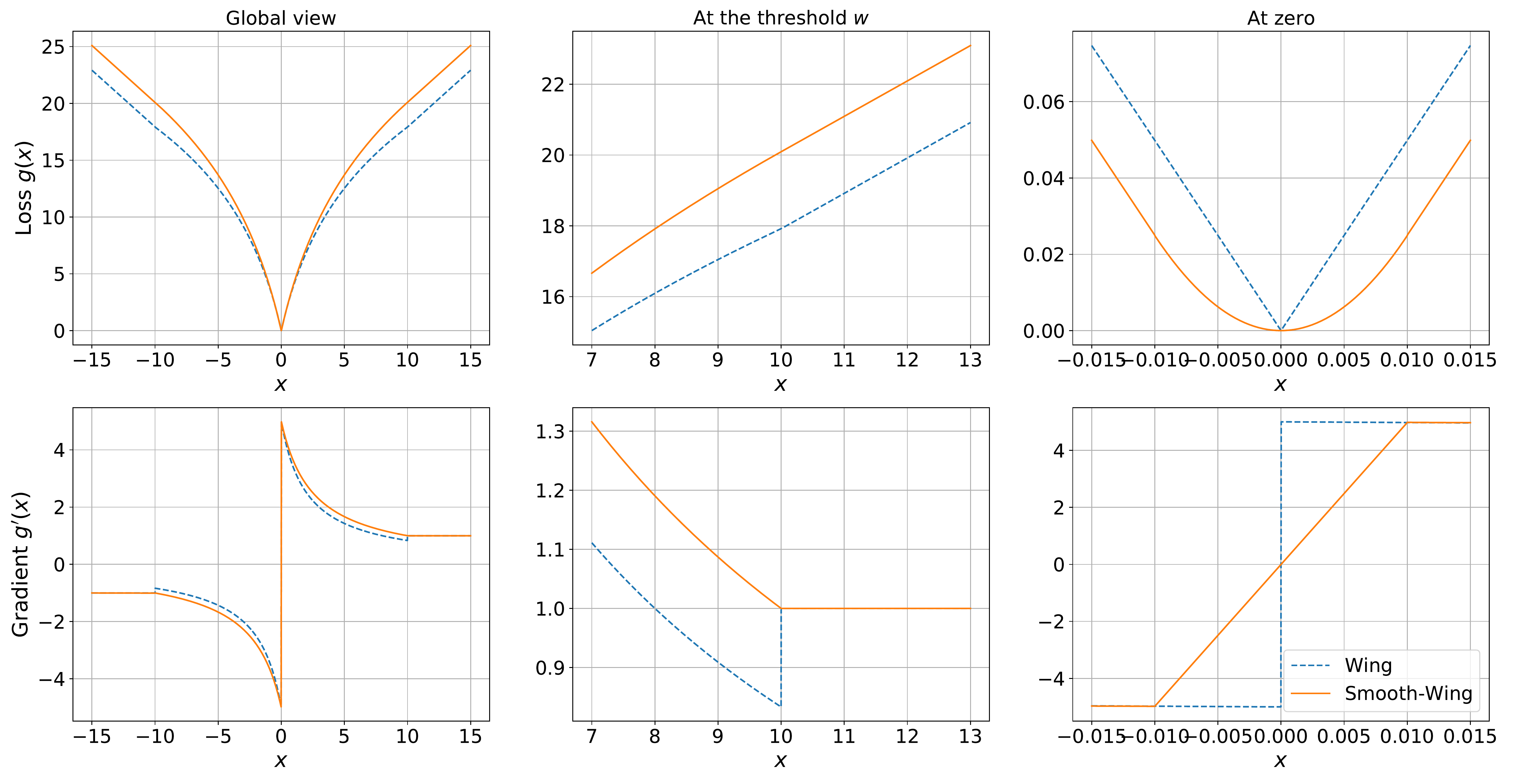}
    \caption{Comparison of Wing loss and smooth-Wing loss (top) and their gradient (bottom) in the global view (left), at the outer threshold $w$ (middle), and at $x$ equals zero (right). The parameters are set as follows: $w$ = 10, $\epsilon$ = 2, and only for smooth-Wing, $t$ = 0.01. 
    For the Wing loss (blue dashed lines), the gradient changes abruptly at the points $|x| = w$ (bottom-middle) and at $x = 0$ (bottom-right).
    On the other hand, the proposed smooth-Wing loss (orange solid lines) is designed to eliminate these gradient discontinuities. 
    }
    \label{fig:wing_vs_s-wing}
\end{figure*}
Several loss functions have commonly been used to train landmark localization models, including L1 loss ($L_1(x) = |x|$), and L2 loss ($L_2(x) = \frac{1}{2} x^2$), and the smooth-L1 loss \cite{girshick2015}, which is defined as:
\begin{equation}
    \label{eq:smooth-l1}
    \text{smooth-}L_1(x) =
    \begin{cases}
        \ \frac{1}{2} x^2     & \text{if } |x| < 1 \\
        \ |x| - \frac{1}{2}   & \text{otherwise.} 
    \end{cases}
\end{equation}
Feng \textit{et al.} \cite{Feng2017} reported another loss function---Wing loss--- which is superior to other loss functions for landmark localization tasks.
The key idea in the Wing loss is to force the model to pay more attention to small errors to improve the accuracy of the predicted landmarks. 
When prediction error is larger than a positive threshold, the Wing loss behaves like L1; otherwise, a logarithm function is used to compute the loss. 
In \cite{Feng2017}, the Wing loss is defined as follows:
\begin{equation}    
    \label{eq:wing}
    \text{wing}(x) = 
    \begin{cases}
        \ w \text{ ln}(1 + \frac{|x|}{\epsilon})     & \text{if } |x| < w \\
        \ |x| - c   & \text{otherwise.} 
    \end{cases}
\end{equation}
where $w$ is the threshold, $\epsilon$ is a parameter controlling the steepness of the logarithm part, and $c = w - w \text{ ln}(1 + w/\epsilon)$. 

However, as shown in Fig. \ref{fig:wing_vs_s-wing}, the Wing loss produces the discontinuity of the gradient at the threshold $w$, as pointed out in \cite{Wang2019c}, and at zero error, which might affect the stability of training. 
In this work, we propose a modified Wing loss that is smooth everywhere and investigate its effectiveness.


\section{Proposed methods}\label{sec:proposed}

\subsection{Localization Transformer (LOTR)}\label{subsec:lotr}

This section explains, in detail, the proposed Transformer-based landmark localization network, named Localization Transformer (LOTR). 
Since LOTR is a coordinate regression approach that directly maps an input image into a set of predicted landmarks, it requires neither heatmap representation nor any post-processing step.  
Fig. \ref{fig:lotr} illustrates an overview of the architecture of the proposed LOTR framework.
An LOTR model consists of three main modules, which include 1) a visual backbone, 2) a Transformer network, and 3) a landmark prediction head.

The visual backbone takes an RGB image as input with the aim to capture context and produce a feature map as output.
In this work, we exploit a pre-trained CNN such as MobileNetV2 \cite{sandler2018}, ResNet50 \cite{resnet2016}, or HRNet \cite{Wang2021HRNet} to compute a feature map. 
We apply 1$\times$1 convolution to reduce the channel dimension of the feature map.
Since the resolution of the feature map generated from the CNN backbone might be very low, e.g., 6$\times$6 pixels for a 192$\times$192 input image, we optionally increase the resolution by using upsampling layers such as deconvolution.

We then utilize a Transformer network \cite{Vaswani} to enrich the feature representations while maintaining the global information in the feature map. 
As shown in Fig. \ref{fig:lotr}, the Transformer module is composed of a Transformer encoder and a Transformer decoder. 
Since Transformers were designed to process sequential data \cite{Vaswani}, we convert the feature map $\mathbf{F} \in \mathbb{R}^{W \times H \times C}$, obtained from the visual backbone, into a sequence of tokens $\mathbf{X}^0 \in \mathbb{R}^{WH \times D}$.
As shown in \eqref{eq:F2X}, we use a $1 \times 1$ convolution layer to reduce the channel dimension $C$ of each pixel in $\mathbf{F}$ to a smaller dimension $D \leq C$, followed by reshaping into a sequence of tokens:
\begin{equation}
\label{eq:F2X}
    \mathbf{X}^0 = \text{reshape}(\text{conv1}{\times}\text{1}(\mathbf{F})).
\end{equation}

A Transformer encoder, i.e., a stack of $L$ encoder layers, receives this sequence of tokens as input.
Each encoder layer consists of two sublayers: 1) Multi-Head Self Attention (MHSA) and 2) PFFN.
Both these sublayers have residual connections and layer normalization applied to them.
As mentioned in Section \ref{subsec:detr}, the MHSA, which is a special type of MHA that establishes the relationship between tokens in the input sequence by computing the attention by linearly projecting each token into query, key, and value vectors and subsequently using the query and key vectors to calculate the attention weight applied to the value vectors. 
The output from the MHSA sublayer (the same size as its input) is then fed into PFFN to further transform the input sequence's representation. 
Similar to DETR \cite{Carion2020}, 2D-positional encoding is added only to the query and key in each encoder layer. 
The $l$-th encoder layer is defined as follows:
\begin{equation}
     \mathbf{X}^{l} = \text{enc}^l(\mathbf{X}^{l-1}, \mathbf{P}),
\end{equation}
\begin{equation}
    \text{enc}^l(\mathbf{X}, \mathbf{P}) = \text{enc}^l_2(\text{enc}^l_1(\mathbf{X}, \mathbf{P})),
\end{equation}
\begin{equation}
    \text{enc}^l_1(\mathbf{X}, \mathbf{P}) = \text{LN}(\text{MHA}^l(\mathbf{X} + \mathbf{P}, \mathbf{X} + \mathbf{P}, \mathbf{X}) + \mathbf{X}),
\end{equation}
\begin{equation}
    \text{enc}^l_2(\mathbf{X}) = \text{LN}(\text{PFFN}^l(\mathbf{X}) + \mathbf{X}),
\end{equation}
\begin{equation}
    \text{PFFN}^l(\mathbf{X}) = \text{ReLU}(\mathbf{X}\mathbf{w}^l + \mathbf{b}^l),
\end{equation}
where $\mathbf{X}^{l-1}$ and $\mathbf{X}^{l}$ denote the input and output of the $l$-th layer, respectively; $\mathbf{P} \in \mathbb{R}^{WH \times D}$ is the trainable 2D-positional encoding; $\text{LN}$ denotes layer normalization; while $\mathbf{w}^l$ and $\mathbf{b}^l$ are the weight and bias of the PFFN layer, respectively.
These processes repeat $L$ times using the output from the previous encoder layers as input. 
The output of the Transformer encoder is a transformed sequence $\mathbf{X}^L \in \mathbb{R}^{WH \times D}$, the same dimension as its input.

Following this encoding operation, the output from the Transformer encoder is then fed into a Transformer decoder, i.e., a stack of $L$ decoder layers. 
Each decoder layer consists of three sublayers: 1) MHSA, 2) Multi-Head Cross-Attention (MHCA), and 3) PFFN. 
The first and the third are similar to those of the encoder layers. 
However, the input to the first sublayer of the first decoder layer is a sequence of landmark queries $\mathbf{Y}^0 \in \mathbb{R}^{N \times D}$; each is an embedding of the same dimension $D$. 
The number of landmark queries equals $N$, which is the number of landmarks to predict. 
In this work, landmark queries are learnable parameters of LOTR, which are optimized during model training. 
The second sublayer, i.e., MHCA, takes the output of the first sublayer (MHSA) and the output generated by the encoder, i.e., $\mathbf{X}^L$, as inputs, and then computes the relationship between tokens in both sequences. 
The third sublayer, i.e., PFFN, then processes the output from the second sublayer.
Like the Transformer encoder, all three of these sublayers have residual connections and layer normalization applied to them. 
The $l$-th decoder layer is defined as follows:
\begin{equation}
     \mathbf{Y}^{l} = \text{dec}^l(\mathbf{Y}^{l-1}, \mathbf{Y}^0, \mathbf{X}^L, \mathbf{P}),
\end{equation}
\begin{equation}
\begin{split}
    \text{dec}^l(\mathbf{Y}, &\mathbf{Y}^0, \mathbf{X}^L, \mathbf{P}) \\
    &  = \text{dec}^l_3(\text{dec}^l_2(\text{dec}^l_1(\mathbf{Y}, \mathbf{Y}^0), \mathbf{X}^L, \mathbf{P})),
\end{split}
\end{equation}
\begin{equation}
\begin{split}
    \text{dec}^l_1(&\mathbf{Y}, \mathbf{Y}^0) \\ 
    & = \text{LN}(\text{MHA}^l_1(\mathbf{Y} + \mathbf{Y}^0, \mathbf{Y} + \mathbf{Y}^0, \mathbf{Y}) + \mathbf{Y}),
\end{split}
\end{equation}
\begin{equation}
\begin{split}
    \text{dec}^l_2(&\mathbf{Y}, \mathbf{X}^L, \mathbf{P}) \\ 
    & = \text{LN}(\text{MHA}^l_2(\mathbf{Y} + \mathbf{Y}^0, \mathbf{X}^L + \mathbf{P}, \mathbf{X}^L) + \mathbf{Y}),
\end{split}
\end{equation}
\begin{equation}
    \text{dec}^l_3(\mathbf{Y}) = \text{LN}(\text{PFFN}^l(\mathbf{Y}) + \mathbf{Y}),
\end{equation}
where $\mathbf{Y}^{l-1}$ and $\mathbf{Y}^{l}$ denote the input and output of the $l$-th layer, respectively.
The Transformer decoder produces $\mathbf{Y}^L \in \mathbb{R}^{N \times D}$ as output.

The landmark prediction head takes as input the sequence $\mathbf{Y}^L$ and outputs $\hat{\mathbf{Z}} \in \mathbb{R}^{N \times 2}$, which stores the predicted $x$- and $y$-coordinates of the $N$ landmarks.
This work exploits a simple PFFN with two hidden layers with ReLU activation. 
The output layer, however, which consists of only two nodes, is without any activation function. 
In particular, the predicted landmarks $\hat{\mathbf{Z}}$ are computed as follows:
\begin{equation}
    \hat{\mathbf{Z}} =  \text{ReLU}(\mathbf{Y}^L\mathbf{w}_1^z + \mathbf{b}_1^z )\mathbf{w}_2^z + \mathbf{b}_2^z,
\end{equation}
where $\mathbf{w}_j^z$ and $\mathbf{b}_j^z$ are the weight and bias of the $j$-th layer ($j \in \{1, 2\}$), respectively.

Similar to \cite{Carion2020}, the computational complexity of the MHSA in an encoder layer grows quadratically with the feature map size, i.e., $O((WH)^2D + WHD^2)$, while that in a decoder layer depends on the number of landmarks instead, i.e., $O(N^2D + ND^2)$. On the other hand, the complexity of MHCA in a decoder layer grows linearly with respect to the feature map size and the number of landmarks: $O(NWHD + (N+WH)D^2)$.

\subsection{Smooth-Wing loss}\label{subsec:loss}

Since the proposed LOTR predicts a fixed number of landmarks ($N$), a more complicated Hungarian loss used in DETR \cite{Carion2020} is not required. 
The proposed LOTR can be trained end-to-end with a standard loss function given cropped-and-aligned face images as input.
In particular, during the training phase, the predicted landmarks $\hat{\mathbf{Z}}$ generated by LOTR are compared with the ground truth landmarks $\mathbf{Z} \in \mathbb{R}^{N \times 2}$ to compute the loss, which is defined as:
\begin{equation}
    \label{eq:loss}
    \text{loss}(\mathbf{Z}, \hat{\mathbf{Z}}) = \sum_{i=1}^N \sum_{j=1}^2 g(z_{ij} - \hat{z}_{ij}),
\end{equation}
where the loss function $g(x)$ can be any standard loss function such as L1, L2, smooth-L1, or Wing loss, which are described in Section \ref{subsec:wing}. 
Although Feng \textit{et al.} \cite{Feng2017} reported that the Wing loss was superior to other loss functions for landmark localization, its major drawback is the gradient discontinuity at the threshold $w$ and around the zero error (Fig. \ref{fig:wing_vs_s-wing}). 
This discontinuity can affect the convergence rate and the stability of training.

In this work, we also propose a modified Wing loss, named smooth-Wing loss ($\text{s-wing}(x)$), which is given by:
\begin{equation}
    \label{eq:smooth-wing}
    \text{s-wing}(x) =
    \begin{cases}
        \ s x^2     & \text{if } |x| < t \\
        \ |x| - c_1 - c_2   & \text{if } |x| > w \\
        \ (w + \epsilon) \text{ ln}(1 + \frac{|x|}{\epsilon}) - c_2     & \text{otherwise},
    \end{cases}
\end{equation}
\begin{equation}
    s = \frac{w + \epsilon}{2t(\epsilon + t)},
\end{equation}
\begin{equation}
    c_1 = w - (w + \epsilon) \text{ ln}(1 + \frac{w}{\epsilon}),
\end{equation}
\begin{equation}
    c_2 = s t^2,
\end{equation}
where $t$ is an additional threshold ($0 < t < w$).
When the error is smaller than the inner threshold $t$, it behaves like L2 loss, allowing the gradient to be smooth at zero error; otherwise, it behaves like the Wing loss. 
We define the constants $s$, $c_1$, and $c_2$ to smoothen the loss at the inner threshold $t$ as well as at the outer threshold $w$. 
As shown in Fig. \ref{fig:wing_vs_s-wing} (bottom-right), the gradient of the smooth-Wing loss changes linearly when the absolute error $|x|$ is smaller than the inner threshold $t$. 
Moreover, the gradient discontinuities at $|x| = w$ are also eliminated, as shown in Fig. \ref{fig:wing_vs_s-wing} (bottom-middle).


\section{Experiments}\label{sec:exp}


\subsection{Datasets}\label{subsec:datasets}

We conducted experiments to measure the performance of the proposed LOTR models on two benchmark datasets: 1) the 106-point JD landmark dataset \cite{liu2019} and 2) the Wider Facial Landmarks in-the-Wild (WFLW) dataset \cite{Wu2018}.

The JD-landmark dataset contains images from other face landmark datasets including as 300W \cite{Sagonas2013, Sagonas2016}, LFPW \cite{Belhumeur2013}, AFW \cite{Zhu2012}, HELEN \cite{brandt2012}, and IBUG \cite{sagonas2013b}, covering large variation of face pose and expression. 
These images are re-annotated with 106-point landmarks, which provide more information about the face structure than any other face landmark dataset.  
The JD dataset consists of 11,393 images for training and 2,000 images each for validation and testing. 

The WFLW dataset, which is based on a well-known face detection dataset called WIDER FACE \cite{yang2016wider}, was recently proposed to be a new benchmark for facial landmark localization.
The WFLW dataset consists of 10,000 faces: 7,500 for training and 2,500 as test images. 
Compared to other previous datasets, the WFLW dataset is manually annotated with 98 landmarks for each face, providing more information on the face structure. 
It is an extremely challenging dataset due to large pose, expression, and occlusion variations.
It also supplies six other annotations of face properties, including pose, expression, illumination, make-up, occlusion, and blur. 
More than 78\% of images in the provided test set have annotation with one or more, up to four, properties.

Following \cite{earp2021sub}, we used a ResNet50-based face detector, proposed by Deng \textit{et al.} \cite{deng2019}, in our pre-processing step. 
In particular, we used the bounding box and a set of five simple landmarks (i.e., eye centers, nose tip, and mouth corners) obtained from the detector to crop and align the detected faces.
We then resized each input image to 192$\times$192 pixels for the JD-landmark dataset and to 256$\times$256 pixels for the WFLW dataset before feeding it to our model.
The detail of the pre-processing step is described in \cite{earp2021sub}.

\subsection{Evaluation metrics}\label{subsec:metrics}

Following \cite{Wang2019c, Kumar2020, xiong2020, earp2021sub}, we used the standard metrics including the normalized mean error (NME), the failure rate, and the area under the curve (AUC) of the cumulative distribution to evaluate and compare landmark localization algorithms.
The NME is computed across all predicted landmarks as follows:
\begin{equation}
    \label{eq:nme}
    \text{NME} = \frac{1}{N} \sum_{i=1}^N \frac{||\mathbf{z}_i - \hat{\mathbf{z}}_i||_2} {d},
\end{equation}
where $N$ is the number of landmarks, $\mathbf{z}_i$ and $\hat{\mathbf{z}}_i$ denotes the $i$-th ground truth landmark and the $i$-th predicted landmark, respectively, and $d$ is a normalization factor. 
For the JD-landmark dataset, following \cite{liu2019}, the normalization factor $d$ was defined as $\sqrt{W_\text{bbox} H_\text{bbox}}$, where $W_\text{bbox}$ and $H_\text{bbox}$ are the width and height of the bounding box enclosing all the ground truth landmarks, respectively. 
While, for the WFLW dataset, the inter-ocular distance, i.e., the distance between the outer eye corners, was used as the normalization factor.
If the NME of a test image is above a threshold, it is considered a failure.
The failure rate is, therefore, the rate of failure cases. 
The threshold was set to 8\% for the JD-landmark dataset and 10\% for the WFLW dataset.
The AUC is computed from the cumulative error distribution, representing the proportion of images with NME smaller than the threshold.
Therefore, a larger AUC represents a more accurate algorithm. 


\subsection{Model training}\label{subsec:training}

Table \ref{tab:lotr_models} presents the configuration of three LOTR models, namely LOTR-M, LOTR-M+, and LOTR-R+, used in the experiments with the JD-landmark dataset. 
The LOTR-M model is the base model that utilizes a MobileNetV2$_{1.0}$, pre-trained on the ImageNet dataset \cite{imagenet09}, as the visual backbone, generating a feature map of size $6\times6\times1280$ from a $192\times192$ RGB image. 
We reduced the number of feature channels to 64 by a 1$\times$1 convolution, then reshaping into a sequence of tokens.
Subsequently, we used a Transformer module with two encoder layers and two decoder layers to process the sequence, followed by a dropout technique \cite{srivastava14a} with a dropout rate of 0.1.
Following, we used a landmark prediction head, i.e., a PFFN consisting of two hidden layers with 512 nodes, with ReLU activation, followed by an output layer with two nodes, to process each token in the output sequence of the Transformer.
In the LOTR-M+ model, to increase the resolution of the feature map, we inserted upsampling layers between the MobileNetV2$_{1.0}$ backbone and the Transformer module.
In particular, we utilized a 1$\times$1 convolution to reduce the number of channels to 256 and then applied two deconvolution layers (128 and 64 filters, with filter size  = $4\times4$) to increase the feature resolution to $24\times24$ pixels before feeding into the Transformer.
The LOTR-R+ model adopts the same configuration as LOTR-M+, except the backbone is changed to a pre-trained ResNet50.
For the WFLW dataset, we experimented with another model, named LOTR-HR+, that uses the same architecture as LOTR-M+ and LOTR-R+ but exploits HRNet \cite{Wang2021HRNet} as the backbone.

\begin{table}
\caption{The architectures of LOTRs used in the experiments with the JD-landmark dataset.}
\centering
\begin{tabular}{lccc}
\hline
                 & LOTR-M & LOTR-M+ & LOTR-R+ \\
\hline
Backbone         & MNetV2$_{1.0}$ & MNetV2$_{1.0}$ & ResNet50 \\
1$\times$1 conv. & 64 & 256 & 256 \\
Upsampling       & - & 128, 64 & 128, 64 \\
\hline
Transformer      & 2, 2 & 2, 2 & 2, 2 \\
\hline
PFFN             & 512, 512, 2 & 512, 512, 2 & 512, 512, 2 \\
\hline
\end{tabular}
\newline
\label{tab:lotr_models}
\end{table}

We initialized the parameters of the LOTR models using He's method \cite{he2015}. 
We used a standard normal distribution to initialize the positional encoding and the landmark queries, both learnable.
For the landmark queries, we specifically used a standard deviation of $10^{-4}$ for the initialization.
We trained the LOTR models using the LAMB optimizer \cite{You2019a} with a base learning of $10^{-3}$ for 100 epochs while reducing the learning rate with a factor of 0.1 at epochs $50$ and $75$ and setting the batch size to $32$.
We used the smooth-Wing loss, described in Section \ref{subsec:loss}, setting the inner threshold ($t$) to 0.01, the outer threshold ($w$) to 10, and the steepness control parameter ($\epsilon$) to 2
as the parameters.
For data augmentation and training tricks, we used the same steps described by \cite{earp2021sub}.
We implemented the model using MXNet framework \cite{chen2015} with Gluon libraries \cite{guo2020}, using a single NVIDIA Titan X GPU for training.

\subsection{Results on the WFLW dataset}\label{subsubsec:wflw}

In this section, we compare the proposed LOTR with several state-of-the-art methods, including Look-at-Boundary (LAB) \cite{Wu2018}, Wing loss \cite{Feng2017}, adaptive Wing loss (AWing) \cite{Wang2019c}, LUVLi \cite{Kumar2020}, Gaussian vector (GV) \cite{xiong2020}, and Heatmap-In-Heatmap (HIH) \cite{lan2021hih}. As shown in Table \ref{tab:wflw}, our proposed LOTR-HR+ achieves an NME of 4.31\%, clearly outperforming LAB, Wing, AWing, and LUVLi methods, and yields an AUC of 60.14\%, surpassing all state-of-the-arts by a large margin (0.44--6.91 points).
While comparable with GV in terms of NME and failure rate, our LOTR model achieves a better AUC and does not require any post-processing step.
Although our proposed LOTR model does not surpass the performance of HIH in terms of NME and failure rate, it outperforms this state-of-the-art model on the AUC metric.

\begin{table*}
\caption{Comparison with the state-of-the-arts on the WFLW dataset.}
\centering
\begin{tabular}{llrrrrrrr}
\hline
Metric        & Method             & Test set & \multicolumn{6}{c}{Subsets}  \\
                                              \cline{4-9}
              &                    &          & Pose  & Expression & Illumination & Make-up & Occlusion & Blur  \\
\hline
NME (\%)      & LAB                &  5.27    & 10.24 &  5.51      &  5.23        &  5.15   &  6.79     &  6.32 \\
              & Wing               &  5.11    &  8.75 &  5.36      &  4.93        &  5.41   &  6.37     &  5.81 \\
              & AWing              &  4.36    &  7.38 &  4.58      &  4.32        &  4.27   &  5.19     &  4.96 \\
              & LUVLi              &  4.37    &  -    &  -         &  -           &  -      &  -        &  -    \\
              & GV                 &  4.33    &  7.41 &  4.51      &  4.24        &  4.18   &  5.19     &  4.93 \\
              & HIH                &  4.18    &  7.20 &  4.19      &  4.45        &  3.97   &  5.00     &  4.81 \\
              \cline{2-9}
              & LOTR-HR+           &  4.31    &  7.50 &  4.62      &  4.21        &  4.16   &  5.22     &  4.89 \\
\hline
Failure       & LAB                &  7.56    & 28.83 &  6.37      &  6.73        &  7.77   & 13.72     & 10.74 \\
rate (\%)     & Wing               &  6.00    & 22.70 &  4.78      &  4.30        &  7.77   & 12.50     &  7.76 \\
              & AWing              &  2.84    & 13.50 &  2.23      &  2.58        &  2.91   &  5.98     &  3.75 \\
              & LUVLi              &  3.12    &  -    &  -         &  -           &  -      &  -        &  -    \\
              & GV                 &  3.52    & 16.26 &  2.55      &  3.30        &  3.40   &  6.79     &  5.05 \\
              & HIH                &  2.96    & 15.03 &  1.59      &  2.58        &  1.46   &  6.11     &  3.49 \\
              \cline{2-9}
              & LOTR-HR+           &  3.52    & 16.31 &  3.53      &  3.16        &  2.43   &  7.07     &  5.19 \\
\hline
AUC (\%)      & LAB                & 53.23    & 23.45 & 49.51      & 54.33        & 53.94   & 44.90     & 46.30 \\
              & Wing               & 55.04    & 31.00 & 49.59      & 54.08        & 55.82   & 48.85     & 49.18 \\
              & AWing              & 57.19    & 31.20 & 51.49      & 57.77        & 57.15   & 50.22     & 51.20 \\
              & LUVLi              & 57.70    &  -    &  -         &  -           &  -      &  -        &  -    \\
              & GV                 & 57.75    & 31.66 & 56.36      & 58.63        & 58.81   & 50.35     & 52.42 \\
              & HIH                & 59.70    & 34.20 & 59.00      & 60.60        & 60.40   & 52.70     & 54.90 \\
              \cline{2-9}
              & LOTR-HR+           & 60.14    & 36.59 & 57.23      & 60.92        & 60.70   & 54.65     & 55.53 \\
\hline
\end{tabular}
\par
\label{tab:wflw}
\end{table*}    

\begin{figure*}
    \centering
    \includegraphics[width=0.8\textwidth]{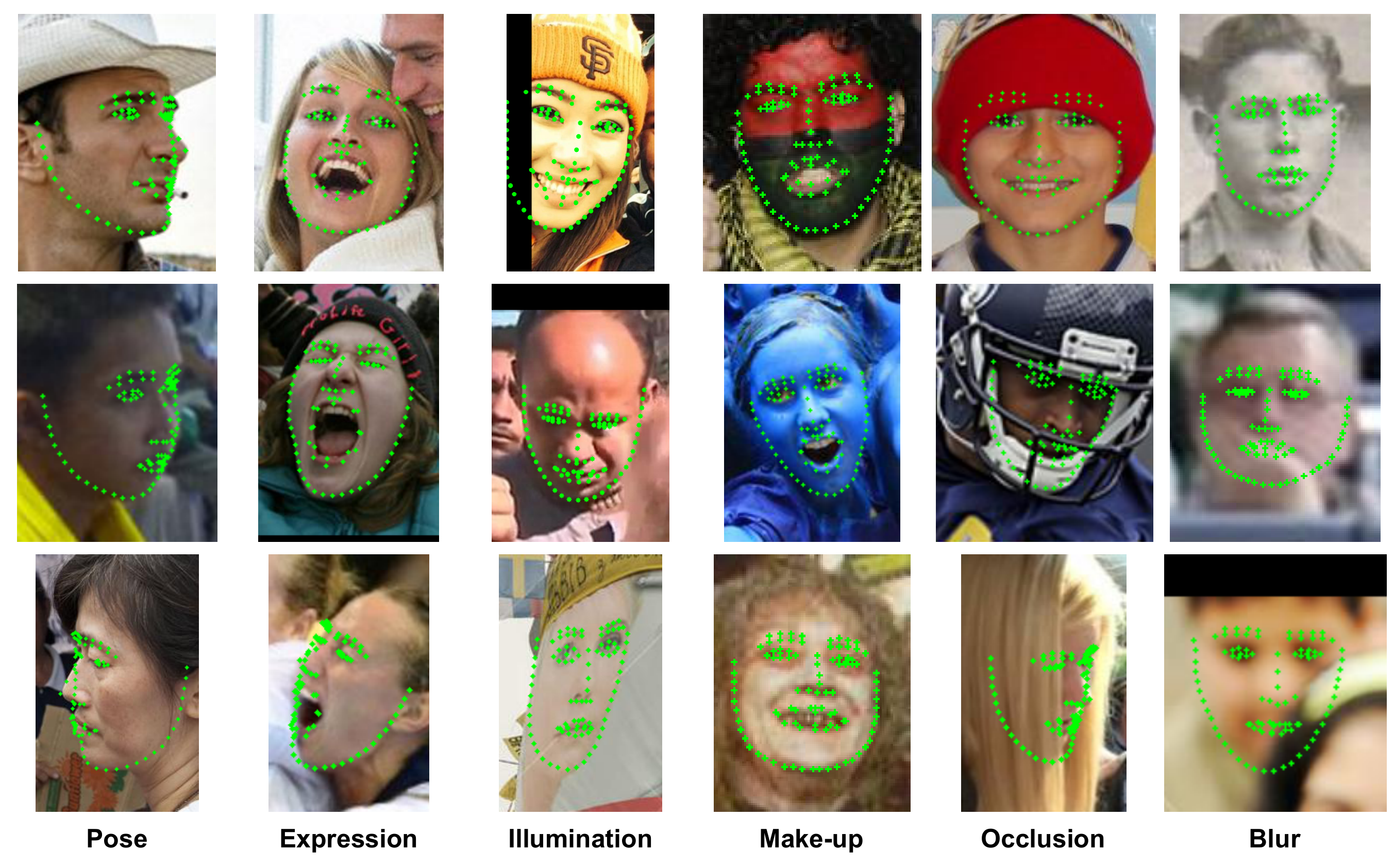}
    \caption{Sample images of the test set of the WFLW dataset with predicted landmarks from our model. Each column displays the images with different subsets. Each row displays images with a different range of NMEs: < 0.05 (top), 0.05--0.06 (middle), and > 0.06 (bottom).}
    \label{fig:output}
\end{figure*}

Fig. \ref{fig:output} presents sample images of each subset from the test set of the WFLW dataset by visualizing predicted landmarks overlaid on the input images.
In this figure, we show images with different ranges of the NME where the first row has NME less than 0.05; the middle row has NME in the range of 0.05 to 0.06, and the last row has NME more than 0.06.
The figure shows the face landmark localization ability of our model, which can accurately locate facial key points with different variations, i.e., large pose, exaggerated expression, extreme illumination, make-up, occlusion, and blur.

\subsection{Results on the JD-landmark dataset}\label{subsubsec:jd}

\begin{table*}
\caption{The evaluation results for different LOTR models on the JD-landmark test set; $\dagger$ and $\ddagger$ denote the first and second place entries.}
\centering
\begin{tabular}{lccc}
\hline
Model              & AUC (\%) & Failure rate (\%) & NME (\%) \\

\hline
LOTR-R+            & 87.71 & 0.00 & 0.98 \\
LOTR-M+            & 87.21 & 0.00 & 1.02 \\
Earp \textit{et al.} (ResNet50) \cite{earp2021sub}         & 87.06 & 0.00 & 1.03 \\
LOTR-M             & 87.05 & 0.00 &  1.03\\   
Earp \textit{et al.} (MobileNetV2$_{1.0}$) \cite{earp2021sub}    & 86.49 & 0.00 & 1.08 \\
$\dagger$Baidu-VIS & 84.01 & 0.10 & 1.31 \\
Xiong \textit{et al.} \cite{xiong2020}   & 83.34 & 0.10 & 1.35 \\
$\ddagger$USTC     & 82.68 & 0.05 & 1.41 \\
\hline
\end{tabular}
\par
\label{tab:jd}
\end{table*}    

Table \ref{tab:jd} presents the performance of the proposed LOTR models, evaluated on the test set of the first Grand Challenge of 106-Point Facial Landmark Localization
\footnote{\label{footnote:jd1}\href{https://facial-landmarks-localization-challenge.github.io}{https://facial-landmarks-localization-challenge.github.io}}. 
For brevity, the table includes only the result of the top two ranked algorithms out of 21 algorithms submitted to the challenge.

Table \ref{tab:jd} shows that the proposed LOTR models, including the smallest model, i.e., LOTR-M, gain more than 3 points in the AUC in comparison to the top two ranked algorithms on the first challenge leaderboard.
The table also compares LOTR models with two recent methods based on the heatmap regression, i.e., \cite{earp2021sub} and \cite{xiong2020}. 
All of the proposed LOTR models surpass the heatmap approach in \cite{xiong2020} by a significant amount of the AUC gain (3.7--4.5 points). 
In comparison with \cite{earp2021sub}, our LOTR models achieve better performance using the same backbone. 
While our smallest model, i.e., LOTR-M, is comparable with the ResNet50 model by \cite{earp2021sub}, our bigger models, i.e., LOTR-M+ and LOTR-R+, surpass it by an AUC gain of 0.15--0.65 points, showing the superiority of our approach over theirs.
Similar to \cite{earp2021sub}, we also utilized a flipping technique during inference to improve the prediction accuracy. 
In particular, we fed the original cropped-and-aligned face images and their horizontally flipped version into the model and averaged their corresponding predicted landmark coordinates.

In terms of floating point operations per second  (FLOPS), the LOTR-M model uses only 0.23 GFLOPS. 
In contrast, the LOTR-M+ model has 47 percent higher GFLOPS than the former due to the presence of upsampling layers; however, operating with only 0.44 GFLOPS. 
The LOTR-R+ model uses 3.23 GFLOPS which is significantly higher than both the LOTR-M and LOTR-M+ models due to its heavier ResNet50 backbone.


\subsection{Effect on face recognition performance}\label{subsec:face_rec}
In recent years, face recognition systems have evolved to contain four modules: face detection, alignment, embedding, and distance computation.
Several existing face detectors predict the location of five naive landmarks that correspond to the center of each eye, the tip of the nose, and the corners of the mouth.
These five landmarks are used for face alignment in accordance to the conventional five-landmark alignment protocol (e.g., \cite{wolf2009,schroff2015,LiuWeiyang2017,wang2018,deng2018,an2020}).
In this section, we investigate how the face alignment process using our proposed LOTR models affects the performance of the state-of-the-art face recognition system.

In our experiments, we first detected the face and then used our LOTR models to obtain the 106 facial landmarks.
Following this, we extracted a subset of the five landmarks for face alignment to make a fair comparison.
We experimented with two different face detectors, namely PyramidKey with MobileNetV2$_{1.0}$ backbone proposed by Earp \textit{et al.} \cite{earp2019} and the publicly available RetinaFace with a ResNet50 backbone from Deng \textit{et al.} \cite{deng2019}.
We exploited the pre-trained LResNet100E-IR from InsightFace\footnote{\label{footnote:insightface}\href{https://github.com/deepinsight/insightface}{https://github.com/deepinsight/insightface}} as the face embedding network.
Table \ref{tab:face_rec} shows the performance of the face embedding network with different combinations of the face detectors and face alignment processes on six benchmark datasets, which include LFW  \cite{lfw}, CFP-FP \cite{cfp-fp}, CALFW \cite{calfw}, CPLFW \cite{cplfw}, IJB-B \cite{ijbb}, and IJB-C \cite{ijbc}.
We report the True Acceptance Rate (TAR) @ False Acceptance Rate (FAR) $=10^{-4}$, following the protocol for IJB-B and IJB-C from ArcFace (see \cite{deng2018,an2020} for more details)
which incorporates both detector score and feature normalization.

The results reported in Table \ref{tab:face_rec} indicate no improvement on LFW with either of the face detectors and the LOTR models.
However, with PyramidKey as the face detector, we see an improvement of 1.85 and 2.47 points on CFP-FP and CALFW with the LOTR-M model while improving by 10.83 points on the CPLFW dataset with the LOTR-R+ model, respectively.
On the other hand, with RetinaFace as the face detector, we see a gain of 1.52 points on CFP-FP with LOTR-M+, while improving by 5.7 points with LOTR-R+ on CPLFW.
Similarly, with both the face detectors, we found an improvement of 0.17 points with LOTR-R+ and 0.33 points with the LOTR-M+ model on the IJB-B and IJB-C datasets.
The improvement in the TAR on CPLFW is note-worthy as this dataset include images with large pose variations.
Thereby, the results suggest that the LOTR models are more robust to pose variations.

\begin{table*}
\caption{The performance (TAR @ FAR $=10^{-4}$) of face recognition on several benchmarks.}
\centering
\begin{tabular}{llcccccc}
\hline
Face detector    & Face landmarks    & LFW   & CFP-FP & CALFW & CPLFW & IJB-B & IJB-C   \\
\hline
PyramidKey       & From detector       & \textbf{99.73} & 94.66  & 84.70 & 53.77 & 94.37 & 94.59   \\
\cite{earp2019} & LOTR-M  & \textbf{99.73} & \textbf{96.51}  & \textbf{87.17} & 61.67 & 94.47 & \textbf{94.92}   \\
                 & LOTR-M+ & \textbf{99.73} & 94.46  & 85.37 & 63.13 & 94.49 & 94.88   \\
                 & LOTR-R+ & \textbf{99.73} & 95.28  & 86.43 & \textbf{64.60} & \textbf{94.54} & 94.89   \\
\hline
RetinaFace       & From detector       & \textbf{99.73} & 94.71  & \textbf{87.47} & 57.97 & 94.43 & 94.59   \\
\cite{deng2019} & LOTR-M  & \textbf{99.73} & 94.68  & 86.20 & 62.90 & 94.48 & \textbf{94.79}   \\
                 & LOTR-M+ & \textbf{99.73} & \textbf{96.23}  & 86.00 & 62.90 & 94.48 & 94.77   \\
                 & LOTR-R+ & \textbf{99.73} & 96.06  & 86.37 & \textbf{63.67} & \textbf{94.54} & 94.74   \\
\hline
\end{tabular}
\par
\label{tab:face_rec}
\end{table*}    

\subsection{Ablation studies}\label{subsec:ablation}

This section further studies the proposed methods in various points, including the computational complexity and model size, the effect of the Transformer network, the number of Transformer encoder/decoder layers, and the proposed smooth-Wing loss. 
All experiments in this section were conducted on the JD-landmark dataset.
\subsubsection{Comparison with a heatmap regression approach}\label{subsubsec:time}

\begin{table*}
\caption{Comparison of prediction accuracy, inference time, and model size.}
\centering
\begin{tabular}{llrrrrrrr}
\hline
\multicolumn{1}{c}{Model} &
\multicolumn{1}{c}{Backbone} &
\multicolumn{2}{c}{AUC (\%)} &
\multicolumn{2}{c}{GPU (ms)} &
\multicolumn{1}{c}{Size} &
\multicolumn{1}{c}{Params} \\
\cline{3-6}
& 
&
\multicolumn{1}{c}{No flip} &
\multicolumn{1}{c}{Flip} &
\multicolumn{1}{c}{No flip} &
\multicolumn{1}{c}{Flip} &
\multicolumn{1}{c}{(MB)} &
\multicolumn{1}{c}{($\times10^6$)}\\
\hline
Earp \textit{et al.} \cite{earp2021sub}   & MobileNetV2$_{1.0}$ & 85.24 & 86.49 & 24.74$\pm$2.08 & 48.28$\pm$3.20 & 12.16 &  2.92  \\
                     & ResNet50          & 86.43 & 87.06 & 24.68$\pm$1.51 & 50.99$\pm$3.67 & 94.10 & 24.42 \\  
\hline
Earp \textit{et al.} \cite{earp2021sub} with   & MobileNetV2$_{1.0}$ & 84.31 & 85.51 & 10.20$\pm$1.45 & 18.90$\pm$0.18 & 12.16 & 2.92  \\
vectorization  & ResNet50          & 84.65 & 85.94 & 11.12$\pm$0.27 & 20.02$\pm$0.35 & 94.10 & 24.42  \\  
\hline  
CNN + FFN & MobileNetV2$_{1.0}$ & 85.85 & 86.37 & 5.51$\pm$0.68 & 8.98$\pm$1.95 & 14.71 & 3.89 \\
    & ResNet50            & 86.18 & 86.91 & 6.09$\pm$0.90 & 9.86$\pm$0.96 & 96.09 & 25.24  \\
\hline  
LOTR-M & MobileNetV2$_{1.0}$  & 86.39 & 87.05 & 5.46$\pm$0.95 & 7.68$\pm$1.08  & 10.59 &  2.81 \\  
LOTR-M+ & MobileNetV2$_{1.0}$ & 86.56 & 87.21 & 5.63$\pm$0.52 & 9.30$\pm$2.81  & 14.03 &  3.71 \\ 
LOTR-R+ & ResNet50          & 87.09 & 87.71 & 6.00$\pm$0.84 & 11.55$\pm$1.03 & 95.97 & 25.21 \\ 
\hline
\end{tabular}
\par
\label{tab:TimeAndSize}
\end{table*}

To study the efficiency of the proposed LOTR models, we measured and compared the inference time with a heatmap-based approach. 
We selected the models with the same backbones from Earp \textit{et al.} \cite{earp2021sub} to serve as the baseline since we followed the same pre-processing procedure, visual backbone, and the bag of tricks.
We reported the result in Table \ref{tab:TimeAndSize} where we ran the models on a high computational CPU (Intel Xeon CPU E5-2698 v4) and NVIDIA Tesla V100 SXM2 GPU with 32 GB of RAM.

In the pre-processing step, all the models used the same face detector by Deng \textit{et al.} \cite{deng2019} as mentioned in Section \ref{subsec:datasets} with a processing time of $43.36\pm4.82$ ms.
Table \ref{tab:TimeAndSize} demonstrates the inference time reduction from the baseline on the GPU by $\sim$4--6$\times$, while maintaining a comparable model size and the number of parameters.
Moreover, when considering the model with MobileNetV2$_{1.0}$ as a backbone, the computational time of LOTR-M and LOTR-M+, with or without the flipping technique, is lower than the MobileNetV2$_{1.0}$ model in \cite{earp2021sub} without the flipping technique. 
This phenomenon is consistent with the larger visual backbone as well.
Table \ref{tab:TimeAndSize} demonstrates that the LOTR-R+ model with flipping is $\sim$4.4$\times$ faster than the MobileNetV2$_{1.0}$ model with flipping in \cite{earp2021sub} and $\sim$2.1$\times$ faster than that model without the flipping technique.

This experimentation demonstrates the complexity of the heatmap regression approach, which relies on a complicated post-processing procedure to generate spatial probability maps. 
Furthermore, some complex operations in its post-processing stage could not efficiently utilize GPU acceleration, resulting in high computational time. 
In contrast, the proposed LOTR models use a Transformer to directly regress the coordinates, thereby, enabling the models to reduce the computational time as it avoids complicated post-processing and is capable of utilizing GPU acceleration as the models consist of only simple operations.

Inspired by Xiong \textit{et al.} \cite{xiong2020}, we conducted another experiment incorporating vectorization and the band pooling module with the baseline approach \cite{earp2021sub}.
We analyzed the effect on model performance and post-processing time reduction with vectorization.
While the post-processing time significantly drops when converting heatmaps to vectorized labels and predictions, there is also a drop in performance. 
The results indicate that the proposed LOTR models still outperform the baseline with vectorization in terms of prediction accuracy and inference time.


\subsubsection{Comparison with a Feed-Forward Network architecture}\label{subsubsec:ablation_mlp}

To investigate the significance of a Transformer's ability in processing spatial features from the visual backbone, we compared its performance with a baseline CNN model with a Feed-Forward Network (FFN) replacing the Transformer module.
For a fair comparison, we applied a $1\times1$ convolution layer after retrieving the feature map from the visual backbone to resemble the channel dimension of the LOTR models.
We then used the FFN as a landmark prediction head to output the coordinates of facial landmarks. 
For comparison purposes, we experimented with MobileNetV2$_{1.0}$ and ResNet50 as the visual backbones and adopted the same loss function and training tricks as described in Section \ref{subsec:training}.

According to the results presented in Table \ref{tab:TimeAndSize}, the performance of the CNNs with FFN head is worse than LOTRs for both flip and no flip setups.
This drop in performance might be due to the incapability of the FFN to capture complex relationships between spatial features.

In contrast, the Transformer module incorporates self-attention and cross-attention mechanisms to model this relationship.
Moreover, the use of landmark queries on the decoder side and the positional encoding might be the reason that the LOTRs can encode abstract information of each landmark, resulting in an improved understanding of the models to perform direct regression.
This shows that the Transformer module is more effective than the FFN.


\subsubsection{Comparison of loss functions}\label{subsubsec:ablation_loss}

In this section, we investigated the effect of the proposed smooth-Wing loss, as described in Section \ref{subsec:loss}, by comparing it with standard loss functions (e.g., L1, L2, and smooth-L1) and Wing loss (Section \ref{subsec:wing}).
We conducted the experiment with the same training process described in Section \ref{subsec:training}.
For the Wing loss, we set the threshold $w$ = 10 and $\epsilon$ = 2.

Table \ref{tab:losses} shows the results from different loss functions on the JD-landmark test set \cite{liu2019}.
The results show that LOTRs with L1 achieve comparable performance with smooth-L1, while with L2, the proposed models' performance is worse than other loss functions.
This result also coincides with the results from Feng \textit{et al.} \cite{Feng2017}.
Unlike L1 loss, which maintains a constant gradient value across the error range, L2 produces a smaller gradient near zero, which causes the models to ignore small error values.
Thus, L2 loss is sensitive to outliers making it less responsive to relatively smaller errors.
Consequently, the models trained with L2 loss may end up omitting small errors, which may yield inaccurate predictions.

\begin{table}
\caption{Comparison of prediction accuracy (AUC) with different loss functions.}
\centering
\begin{tabular}{lccc}
\hline
Loss        & LOTR-M & LOTR-M+ & LOTR-R+\\
\hline
L2           & 84.52  & 84.28   & 86.73\\
L1           & 86.90  & 86.95   & 87.44 \\
Smooth-L1    & 86.87  & 86.98   & 87.52 \\
Wing         & 86.92  & 87.04   & 87.68 \\
Smooth-Wing  & \textbf{87.05}  & \textbf{87.21}   & \textbf{87.71} \\
\hline
\end{tabular}
\par
\label{tab:losses}
\end{table}

Moreover, loss functions for landmark localization---Wing and smooth-Wing---consistently outperform standard loss functions.
The focus of these loss functions is on a small error range, which aids in precise landmark coordinate predictions.

Comparing the Wing loss and the proposed smooth-Wing loss, LOTRs trained with the smooth-Wing loss outperform those trained with the Wing loss. 
This result shows the impact of training stability from the smooth-Wing loss.
Fig. \ref{fig:wing_vs_s-wing} shows an uneven gradient of Wing loss at the threshold $w$, which may not be suitable for parameter adjustment.
The proposed smooth-Wing loss smoothens the gradient at the threshold $w$, making training more stable and, thus, essential for parameter adjustment.
Table \ref{tab:jd} and \ref{tab:TimeAndSize} show that with the same visual backbone, our LOTRs with smooth-Wing loss outperform the heatmap-based approach in \cite{earp2021sub}. 
This establishes that training with the smooth-Wing loss helps the proposed LOTRs achieve state-of-the-art performance on the JD-landmark test set.


\subsubsection{The number of Transformer encoder/decoder layers}\label{subsubsec:ablation_tf_layers}

This section studies the impact of layers ($L$) in both the Transformer encoder and decoder.
We experimented with the different values of $L$ while keeping the other hyper-parameters the same.
Table \ref{tab:tf_layer} shows the results from the different numbers of encoder and decoder layers.
The results show that the LOTR models with up to three layers can accurately localize landmarks and yield the highest AUC when $L$ equals 2.
The models become harder to optimize as the number of layers increases from three.
The results of the deepest models, i.e., $L$ equals 6, is NaN, which might be a consequence of training instability when the number of layers becomes very large.

\begin{table}
\caption{Comparison of prediction accuracy (AUC) based on varying number of Transformer layers.}
\centering
\begin{tabular}{cccc}
\hline
Encoder/       & LOTR-M & LOTR-M+ & LOTR-R+ \\
decoder layers &        &         &         \\
\hline
1              &         86.88  &          87.00   &          87.46   \\
2              & \textbf{87.05} &  \textbf{87.21}  &  \textbf{87.71}   \\
3              &         86.86  &          86.92   &          87.42   \\
4              &         44.09  &          85.88   &          60.84   \\
5              &         43.74  &          44.40   &          44.16   \\
6              &         NaN    &          44.14   &          NaN     \\
\hline
\end{tabular}
\par
\label{tab:tf_layer}
\end{table}


\section{Conclusions}\label{sec:conclusion}

We show that our proposed LOTRs outperform other algorithms, including the two current heatmap-based methods on the JD-landmark challenge leaderboard, and are comparable with several state-of-the-art methods on the WFLW dataset.
The results suggest that the Transformer-based direct coordinate regression is a promising approach for robust facial landmark localization.

We evaluate and illustrate that using the LOTRs for face alignment improves the state-of-the-art face recognition performance on the standard benchmark datasets such as CFP-FP, CPLFW, IJB-B, and IJB-C datasets.

We demonstrate how the proposed LOTR outperforms CNN models with FFN, trained under the same conditions. 
The results suggest an effective use of a Transformer network to improve the feature representation from a visual backbone. 
In contrast to other coordinate regression approaches that suffer from spatial information loss, our LOTRs utilize the crucial spatial information for landmark localization tasks.

We also show that the LOTRs are superior to the recently proposed heatmap-based method by Earp \textit{et al.} \cite{earp2021sub} in terms of accuracy and inference time. 
Since the proposed LOTR directly predicts landmark coordinates, it avoids any computationally intensive post-processing required by the heatmap-based method, leading to a $\sim$4--6$\times$ gain in speed during inference. 
Although Xiong \textit{et al.} \cite{xiong2020} exploited their proposed vectorization method to reduce the post-processing time, its downside is the reduction in prediction accuracy.
The end-to-end training behavior of the LOTRs with the smooth-Wing loss also leads to better prediction performance when compared to the heatmap-based methods.

While Feng \textit{et al.} \cite{Feng2017} reported that their proposed Wing loss is superior over other common loss functions such as L2, L1, and smooth-L1, we show that our proposed smooth-Wing loss leads to better optimized models than the Wing loss because of its gradient continuity. 
The results indicate an improvement in training stability and convergence rate using the smooth-Wing loss. 



\section*{Acknowledgment}

We are grateful to our colleagues Christina Kim, Jeff Hnybida, and Justin Cairns for their constructive input and feedback during the writing of this paper. 

\bibliographystyle{IEEEtran}
\bibliography{ms}

\end{document}